# ROBUSTCAPS: A TRANSFORMATION-ROBUST CAPSULE NETWORK FOR IMAGE CLASSIFICATION

### Sai Raam Venkataraman[1], S. Balasubramanian[2], R. Raghunatha Sarma[3]

[1]DMACS, SSSIHL, Puttaparthi, Bengaluru, 515134, Andhra Pradesh, India
[1]vsairaam@sssihl.edu.in,
[2]DMACS, SSSIHL, Puttaparthi, Anantpur, 515134, Andhra Pradesh, India
[2] sbalasubramanian@sssihl.edu.in
[3]DMACS, SSSIHL, Puttaparthi, Anantpur, 515134, Andhra Pradesh, India
[2] rraghunathasarma@sssihl.edu.in

**Abstract**

*Geometric transformations of the training data as well as the test data present challenges to the use of deep neural networks to vision-based learning tasks. In order to address this issue, we present a deep neural network model that exhibits the desirable property of transformation-robustness. Our model, termed RobustCaps, uses group-equivariant convolutions in an improved capsule network model. RobustCaps uses a global context-normalised procedure in its routing algorithm to learn transformation-invariant part-whole relationships within image data. This learning of such relationships allows our model to outperform both capsule and convolutional neural network baselines on transformation-robust classification tasks. Specifically, RobustCaps achieves state-of-the-art accuracies on CIFAR-10, FashionMNIST, and CIFAR-100 when the images in these datasets are subjected to train and test-time rotations and translations.*

## 1. INTRODUCTION

**1.1 Equivariance and geometric transformations**

Convolutional neural networks (CNNs) have, for a decade now, remained as one of the best performing models for computer vision. Their successes in computer vision can be attributed to both advances in computing hardware which allow for deeper models, and to the weight sharing scheme that the correlations in CNNs use.

Under this weight-sharing scheme, pattern detectors are shared translationally so that an object detected at a location may be detected at others even after translations. Figure 1 presents a visual representation of the weight-sharing, and Figure 2 presents a depiction of this property of detection across translations. This property is related directly to the fact that translations of objects in a scene do not affect their categories. Seen more generally, this *translational symmetry* can be generalised to other transformations of the data such as rotations and reflections, or in general, any symmetry transform which usually leaves the category of the data unchanged.

While the detector-sharing of CNNs does not reflect such general symmetries, the simple correlation on grid-locations has been generalised to general groups of symmetry transforms. This generalisation allows for the preservation of transformation symmetries, and results in a greater sharing of detectors in layers. These models, termed group-equivariant CNNs (GCNNs), were first introduced in [1]. The authors extended the correlation in CNNs to groups of more general transformations, such as that formed from the composition of orthogonal rotations, translations, and reflections. Figure 3 shows a visual explanation of equivariance to rotations and translations.

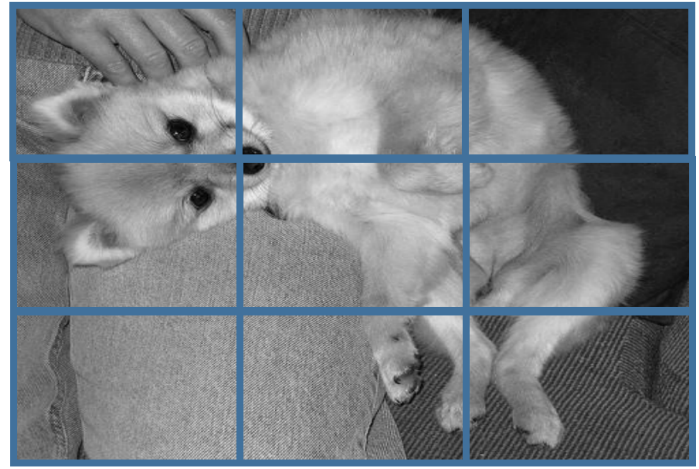

Fig. 1 The above image depicts the usual weight-sharing scheme in CNNs, Each blue rectangle within the image indicates a region where a single correlation with a filter happens. The weights of the CNNs are shared across these rectangles. Thus, for a filter of a CNN-layer, the same pattern is detected at each of the rectangles.
Seen in another manner, it can be said that the filters are translated across the regions of the image. The image is taken from the ImageNet dataset.

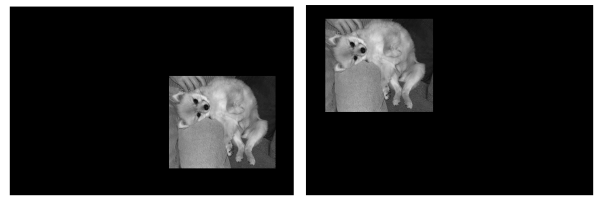

Fig, 2 The image on the left shows a dog on the right-bottom of a black background. The image on the right shows the dog on the top-left of a black background. It can be said that the pattern of the dog has been translated to different locations of the image. A CNN would give the same representation to both the images, only translated to the appropriate locations. The image of the dog is from the ImageNet dataset.

This more general setting in GCNNs allows for greater transformation-robustness that is reflected both in empirical results and in a formal, mathematical guarantee that is termed group-equivariance. Informally, group-equivariance enforces that transformations of the input data be reflected as the same transformations of the output data. Thus, it may be seen as a guarantee of predictability under transformations of the input. Extensions to the original model are a subject of active research and generally involve extending the definition of correlations to various groups [2] [3].

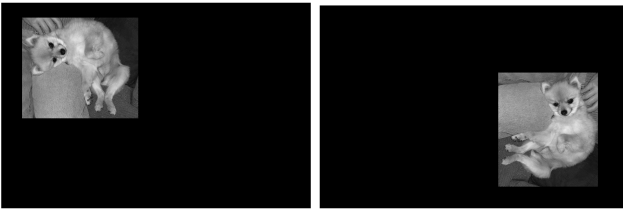

Fig. 3 The image on the left shows a dog on the right-bottom of a black background. The image on the right shows the dog on the top-left of a black background. It can be said that the pattern of the dog has been translated from the top left to the bottom right and then rotated clockwise by 90°. An equivariant CNN would give the same activations to both the images, only translated to the appropriate locations and rotated by the same angle. The image of the dog is from the ImageNet dataset.

## 1.2 Preliminaries: a formal definition of equivariance

Based on the informal description of equivariance that we have provided, we see that equivariance simply means the preservation of a transformation applied on to the input by a neural network. More formally, and generally, equivariance of function is described by the preservation of a group-action on the input space. We present these definitions that were introduced earlier in works such as [1].

**Group:** In order to formally describe the effect of transformations on inputs, it is first necessary to use a structure that can describe the transformations. The concept of a group offers one means of doing so, Groups allow for a conceptual categorisation of geometric transformations such as translations, rotations, shearing, scaling etc. The formal definition is given below.

A set G with a binary operator . defined over G is said to be a group if the following are true.

(*Closure*) For all g, h in G, g.h is also in G.
(*Associativity*) For g, h, k in G, (g.h).k = g.(h.k).
(*Identity*) There exists e in G such that for all g in G, g.e = e.g = g.
(*Inverse*) For all g in G, there exists $g^{-1}$, such that $g.g^{-1} = g^{-1}.g = e$.

A common way to represent geometric transforms is to write them as matrices, and use matrix multiplication as the operator. Thus compositions of transformations can be written as the multiplication of matrices. Examples of these are given in [1].

**Vector space:** With groups serving as a formal structure for transformations, it is also necessary to describe the representation space of neural networks. One choice for this is to use vector spaces to model their activations. Informally, vector spaces are sets that are closed under compositions of addition and scalar multiplication. Examples of vector spaces are $R^n$ using component-wise addition with the set of scalars being real numbers , using multiplication of all components by the same number as scalar multiplication, and matrices that use matrix-addition with real numbers as the set of scalars. Since most neural networks have representations in $R^n$, vector spaces are a natural choice for representing them.

**Group action:** Group actions serve as a connection between groups and vector spaces, allowing for the description of transformations of representations. Formally, they are defined in the following. Consider a group (G, .) and a vector space X. A function f : G × X → X is termed a group action if the following are true.

1. f(e, x) = x, for all x in X, and where e is the identity element of G.
2. f(g, f(h, x)) = f(g.h), for all g, h in G and for all x in X.

In this work, as in [1], we shall consider a specific group action denoted by L. First, consider a group (G, .). Further consider a vector space X. Let f : G → X. Then for all g in G, let the following group action be defined.

$$[L_g f](x) = f(g^{-1} x).$$

f is representative of neural networks. For most CNNs, G is any group that contains the translation group. L describes the remapping of elements from their original locations after a geometric transformation such as translation or rotation.

**Group equivariance:** Given this definition of how transformations affect inputs by remapping the inputs to new locations, equivariance is defined by the preservation of the action of the group on the input space to the output space. Formally, this is defined by the following.

Consider a group (G, .), a vector X, and a function f : G → X. Let T and T' be two group actions defined over G. f is said to be equivariant with respect to T and T' if the following is true for all g in G and x in X.

$$f(T(g, x)) = T'(g, f(x)).$$

**Equivariant convolutions:** CNNs use correlations in their layers that are equivariant to translations. This, generally non-equivariant, operation is described below. We shall see that the equivariant convolution is a generalisation of this operation.

Consider the translation group $G_t$. Consider also a CNN whose l-th layer is a function f : $G_t$ → $R^{d\_l}$, where d_l is a positive integer, and represents the number of channels input to the filters of the l-th layer. For example, for colour images, d_l is 3, where l is the input layer. For deeper layers of a neural network, d_l depends on the neural network model. Let us consider the set of d_(l+1) filters and denote it by F, where each filter is represented by $F^i$ : $G_t$ → $R^{d\_l}$. Moreover, let $f_k(x)$ denote the scalar at the k-th dimension of the d_l dimensional vector that f(x) gives. Similarly, let $F^i_k(x)$ denote the scalar at the the k-th dimension of the d_l dimensional vector that $F^i(x)$ is. The correlation operation between f and $F^i$ is given by:

$$[f * F^i](x) = \sum_{y \text{ in } G_t} \sum_{1 \le k \le d\_(l+1)} (f_k(y) \, F^i_k(y - x)).$$

Given any translation t in $G_t$, translation-equivariance is satisfied and expressed in the following expression. The proof is in [1].

$$[[L_t f] * F^i](x) = [L_t[f * F^i]](x).$$

In simple terms, this means that translating an input and performing correlation gives the same result as correlating and then translating the output. Thus, CNNs are equivariant with respect to translations.

The correlation described above is extended to more general groups by the following definition.

Consider a group G. Like before, consider also a CNN whose l-th layer is a function $f : G \to R^{d\_l}$, where d_l is a positive integer, and represents the number of channels input to the filters of the l-th layer. Let us consider the d_(l+1) filters F, where each filter is represented by $F^i : G_t \to R^{d\_l}$. Moreover, let $f_k(x)$ denote the scalar at the k-th dimension of the d_l dimensional vector that f(x) is. Similarly, let $F^i_k(x)$ denote the scalar at the the k-th dimension of the d_l dimensional vector that $F^i(x)$ is. The group-equivariant correlation is defined as the following.

$$[f * F^i](x) = \sum_{y \text{ in } G} \sum_{1 \leq k \leq d\_(l+1)} (f_k(y) F^i_k(x^{-1}y)).$$

Like the usual correlation, the group-equivariant correlation satisfies a related, but general, condition for equivariance. Given any transformation g in G, the following property is satisfied. The proof is in [1].

$$[[L_g f] * F^i](x) = [L_g[f * F^i]](x).$$

In simpler terms, this means that transforming the inputs and then performing correlation gives the same results as performing correlation and then transforming the output. Note the subtle difference in using $L_t$ and in using $L_g$. The first of these refers to the action of translation, while the second is for a general group. Thus, the equivariance conditions are generalised from equivariance to translations to equivariance to more general transformations.

### 1.3 Limitations of group-equivariant convolutions

GCNNs are transformation-robust by the equivariance guarantee that the correlation operation bestows on each layer. This is because the activations of transformed inputs are only remapped versions of the activations of untransformed inputs. GCNNs have shown, therefore, improved results on transformed data as can be seen in [1], [2], and [3]. However, certain limitations exist.

One limitation that we study and aim to remedy is based on the fact that GCNNs do not learn object-structure. The activations of GCNNs represent the detection of patterns in the inputs. These activations are based only on the activations of shallower layers and the weights of filters. In other words, detection of patterns is based on the existence of shallower patterns [4] [5].

Spatial objects, however, have structure that is reflected in relationships among objects in a visual scene. This is seen in the part-whole relationship among objects and their components, and in the relationships among components. These relationships are transformation-invariant, in that any symmetry transformation of a visual scene preserves them. Figure 4 shows an example of such relationships in images and their transformation-invariance.

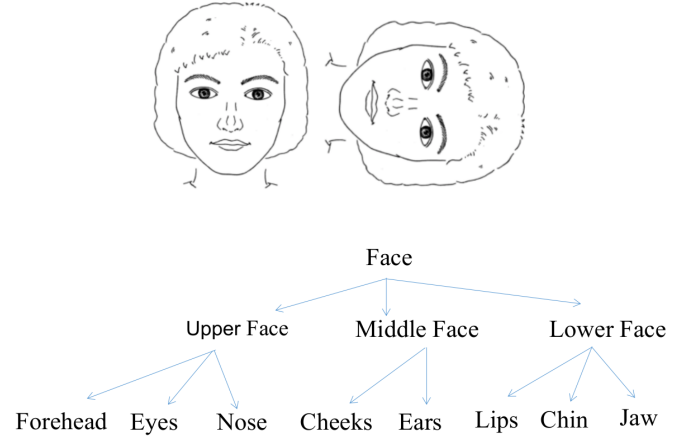

Fig. 4 Consider two images of face - one of them a rotated version of the other. We can still see that components of the face are made up of the same components - just rotated. An example of such a hierarchy is presented in the bottom image. Note that after rotation, the poses of all the parts change; the relations between them do not. More generally, an object such as a face can be hierarchically understood as a combination of parts. Since the composition of a part does not change under transformations such as rotation, the part-whole structure too does not change. Images are from [16].

A neural network that detects objects based on this structure and with the property that the relationships that this structure defines are preserved under transformations would be expected to perform better than GCNNs. This is because better features would be learnt.

### 1.4 Capsule networks, part-whole relationships, and equivariance

The above observation gave rise to the capsule network model [5] [6]. In capsule networks, the vector activations, termed capsules, denote poses of objects as opposed to existential information alone (as in GCNNs). Thus, at each layer of a capsule network, a pattern and its pose, given by the vector capsule, are together detected. In order to do this, unlike GCNNs, capsule networks form deeper capsules from shallower capsules by a specialised procedure termed routing. Routing ensures that the deeper capsules reflect a notion of agreement in the poses of the objects the shallower capsules denote.

The usual method for routing has shallower capsules first pass through prediction sub-networks to give rise to predictions - one for each pair of deeper and shallower capsules. The next step is to combine the predictions for each deeper capsule such that important predictions have a larger weight in the combination. Many algorithms use a weighted-summation to combine predictions, and obtain the weights via iterative procedures [5], [6], graph-based methods [7], or directly from trainable networks [8]. Each of these methods represent a means of finding the extent of agreement among predictions, which is used as a measure of importance.

Why agreement among predictions? Predictions may be thought of as candidate-poses for the deeper capsules. Shallower capsules represent objects that are seen as potential parts of deeper capsules. If an object is part of another object, its prediction for the pose of the bigger object would agree with the predictions for the pose made by other valid parts. This argument is referred to as routing-by-agreement, and forms a key component of capsule network ideas.

The weights for combining the predictions as well as the subnetworks can be considered to represent the part-whole relationships between the objects that the deeper and shallower capsules represent. Any geometric transformation of the inputs must not affect the relationships among capsules. A means of having this property is to specify that the predictions and the routing-weights are both equivariant to transformations of the input. This equivariance in routing can be seen as having invariance in the learnt part-whole relationships as shown in [7].

### 1.5 Our contributions

The above has established the need for models that are both equivariant and are equipped with mechanisms to learn compositional information in the images. However, most capsule models do not satisfy these properties. Formal guarantees for equivariance are not given for most capsule models as other aspects such as routing and an accurate prediction-mechanism are usually formulated.

Models such as SOVNET [7] and group-equivariant capsules (GCAPS) [9] present equivariant capsule models. Both of them, however, have limitations on performance. GCAPS is limited by the fact that its capsules are constrained to be elements of a fixed group. This limitation does not allow it to be highly accurate on data that has significant information that is not easily captured by groups. SOVNET does not have such limitations; nonetheless, its layers are not optimised for achieving state-of-the-art performance on transformation-robust classification on complex data.

Specifically, the architectural aspects of SOVNET need improvement. This improvement can be done by bettering pre-capsule layers to obtain object-centric features for capsules, and also by using less bulky predictors in the capsule layers. A third improvement can be done by adjusting the scales of the predicting vectors in the capsule layers so that improper scaling can be avoided in the weighted-summation of routing.

In light of these observations, we propose a model for capsule networks termed RobustCaps that, like SOVNET, uses GCNNs in a capsule network framework with an equivariant routing procedure. Thus, RobustCaps also displays the property of group-equivariance. Unlike SOVNET, the modules of RobustCaps are designed so as to achieve state-of-the-art results on transformed data classification.

The pre-capsule layers are improved using residual GCNN-layers, while the predictors use a GCNN-layer instead of bulky residual predictors. RobustCaps also uses a global context-normalisation mechanism in its routing mechanism to use appropriate scaling in the predictions.

RobustCaps outperforms several capsule network models on classification of transformed images on CIFAR10, FashionMNIST, and CIFAR100. Further, RobustCaps also outperforms residual networks and group-equivariant residual networks on this task, showing that the learning of part-whole relationships in an equivariant manner is important towards transformation-robust classification.

### 1.6 Implications of our work

Most capsule network models have, until this work, shown lower performance than CNNs and GCNNs. Our work can be seen as an effort to showcase the value of such learning, and could help in the development of better models.

The following summarise our contributions along with their implications:

- We propose **RobustCaps**, a transformation-robust capsule network model. RobustCaps uses a novel routing algorithm along with group-equivariant convolutions to create a capsule network model that displays a high degree of equivariance as well as highly accurate.

- RobustCaps uses a **global-context normalisation** layer with centrality-based routing weights to present an accurate model for capsule networks.

- RobustCaps achieves **state-of-the-art accuracies** on transformed classification on CIFAR-10, FashionMNIST, and CIFAR-100.

- Given the fact that RobustCaps outperforms strong convolutional baselines such as equivariant residual networks, our work can lead to further research on capsule networks.

## 2. Previous work

Research on equivariant convolutions has led to several extensions of the original work in [1]. For example, [3] extends GCNNs to be equivariant to rotations for spherical images. [2] presents a general framework for equivariant CNNs on the euclidean group E(2). Other works can be more theoretical. For example, [10] shows that any linear function that is equivariant can be written as an equivariant convolution. This allows a study of GCNNs to be considered as a study of equivariant models.

Capsule network models aim to remedy the weakness of CNNs and GCNNs that is due to their detection-by-existence procedure for building activations. [4] suggested a remedy by using routing. This was extended to the capsule network models implemented in [5] [6]. Extensions to these models usually improved aspects of capsule networks such as prediction mechanisms.

DeepCaps [11], STAR-CAPS[12], and self-attention capsules [8] are examples of models where different predictors and routing methods were proposed.

DeepCaps uses the dynamic routing of the initial capsule network model in [5], but uses convolutional layers in the initial layers, so as to achieve good accuracies on several datasets. STAR-CAPS and self-attention capsules use the idea of attention-mechanisms in the routing procedure towards the same goal of high performance. The mechanisms used are, however, different.

Models such as group-equivariant capsule networks (GCAPS) [9] and SOVNET [7], on the other hand, recognise that equivariance in routing is necessary towards learning invariant part-whole relationships among capsules. As we point out, these models can show better generalisation for transformations of the data, but are not built so as to achieve the best results. We aim to remedy this.

## 3. Our proposed RobustCaps model

We recognise that in order to achieve state-of-the-art accuracies, the architecture of our model has to be developed along with theoretically pleasing properties such as group-equivariance. To this end, we present a depth-specific construction of components for our model. RobustCaps has four components that correspond to different depths of the model. These are termed as follows: PreCaps, PrimaryCaps, ConvCaps, and ProjCaps. A description of each component is given below. A visual depiction of the model is presented in Figure 6.

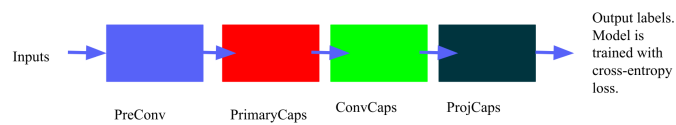

**Figure 5**
A diagram of the modules of RobustCaps.

### 3.1 PreCaps

The PreCaps layer of RobustCaps is a simple residual GCNN network. Specifically, it comprises 7 residual blocks. Each block has the structure as shown in Figure 6. Note that unlike usual residual networks, PreCaps uses group-equivariant correlations.

The use of such a structure for PreCaps has two purposes: obtaining sufficiently useful high-level features for capsules, and maintaining the equivariance of the model for transformation-robustness. The use of convolutional layers before capsule layers is inspired by empirical results in models such as [11]. The intuition behind this is that GCNNs detect sufficiently high-level features that correspond to complex features where the object-based logic of capsule networks and routing can work.

Note that the activations at each channel correspond to the existential information about patterns discovered at a transformational state. This information must now be converted into vectors that can then be used for the pose-centric mechanisms of the capsule layers.

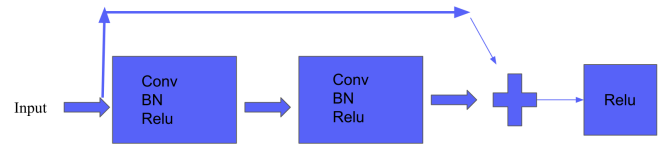

**Figure 6**
The basic residual block for PreCaps. The PreCaps subnetwork is made of 7 such blocks stacked end-to-end.

### 3.2 PrimaryCaps

The PrimaryCaps layer does this by using the scalar activations as inputs for specialised GCNNs to obtain vector capsules. First, as intuition, the PrimaryCaps layer is responsible for detecting the pose and existence of relevant objects of various types. Moreover, since equivariance is desired for transformation-robustness and invariance in the detection of object-relationships in later capsule layers, this layer uses GCNNs for this detection.

Thus, the PrimaryCaps is seen as a set of vector-valued functions defined over a group, where each function is representative of a capsule type of the PrimaryCaps layer. The PrimaryCaps layer is also seen as detecting the poses of an initial set of objects that are then used to detect other patterns in the object-hierarchy of the image,

Thus, each vector-value of a capsule type in the PrimaryCaps layer can be seen as the pose of an instance of an object that the capsule type denotes. Each capsule type in PrimaryCaps is given by a GCNN. Specifically, each dimension of a capsule-type is given by one GCNN filter.

The notion of capsule types and capsules defined on a group extends to other capsule layers. Thus, we explain the distinction between capsules and capsule types using an example. Consider a face detection task. Also consider a capsule network trained for detecting faces. Then, given an image, a capsule layer detects important patterns for the detection task. A capsule type in the capsule layer represents one pattern, such as eyes, ears, or mouths. Depending on the number of instances of these patterns, a capsule type presents vectors at the locations of these patterns. Thus a capsule type that detects a part of a face for this task, would return vectors that capture the generalised pose of all instances of the object. These instances are termed capsules.

Thus, capsule types may be seen as representative of patterns, while capsules are instances of these patterns.

In order to facilitate better training, we normalise the capsules across all types, This is done using layernorm. While layernorm affects exact equivariance, as do the strided convolutions in the residual blocks of PreCaps, it plays a role in improving performance. More generally, multiple 'equivariant' models such as those in [1] and [2] make use of such operations that reduce exact equivariance, but help in empirical accuracy.

### 3.2 ConvCaps

With the poses of object-components detected by the PrimaryCaps, RobustCaps uses a series of ConvCaps layers to sequentially detect objects right up to the class level. It is in these layers that the transformation-invariant relationships among objects are learnt and detected. We describe the details of a ConvCaps layer in the following.

Each ConvCaps layer consists of the following: a GCNN predictor per capsule type, and a global context normalised centrality routing layer. These layers correspond to the steps of prediction and routing-by-agreement that is seen usually in routing algorithms [5] [6] [7] [11].

The inputs themselves are a set of capsule types. The predictions for a capsule type at a layer are formed by performing a correlation of the associated GCNN filter for that type with each input capsule type. The use of GCNNs for prediction of poses in capsules has been done in [7]. However, the use of the GCNNs does not lead to state-of-the-art accuracies. Each predictor there is a residual GCNN subnetwork, and therefore is bulky. In this work, we see that a single GCNN is enough for good performance, showcasing the relative light-weight nature of RobustCaps.

The principle of routing can be seen from the view that a prediction for a deeper capsule that is aligned with other relevant predictions must get a larger weight in the combination of predictions to form the deeper capsule [7]. Based on this principle, a centrality-based routing algorithm was proposed in [7] that combines predictions as a weighted-summation.

Each prediction is considered to be a vertex in a graph, where the edges are weighted by the cosine similarity between the vectors associated with them. Predictions are assigned weights based on the degree-centrality of the vertices. A prediction with a larger degree-centrality displays a greater alignment with the other predictions and therefore obtains a larger weight. Predictions that are not relevant to the compositionality of the input display a low centrality, and are not weighted by as much in the model.

This intuition does not consider the problem of non-uniformity in scale in the predictions. Due to lack of a mechanism to normalise scales of predictions, there could be predictions that are not very well aligned, but with a larger scale that could cause a greater influence in the summation. While the weights themselves would be relatively low, the values of the elements of the vectors could be large enough to cause undue steering to itself.

In order to mitigate this, we introduce a global-context normalising layer prior to routing using degree-centrality. This layer normalises predictions by a shallower capsule type for a deeper capsule type by subtracting the mean and dividing the standard deviation for each element of the capsules across all types and transformational states. In order to incorporate a learnable component into this for better performance, the normalised capsules are multiplied by and divided by two learnable vectors. This layer is easily implemented using a layernorm after the predictions.

After the normalisation, rescaling, and recentering, the centrality of each prediction is calculated using the cosine similarity as an edge-weight. The centralities are softmaxed so as to keep the weights for the predictions for a deeper capsule between 0 and 1. The global-context normaliser is especially useful here, as the softmax layer tends to increase the relative gap between elements. In a setting where non-uniform scales are present, the softmax layer tends to increase the unfairness of weights. However, the use of softmax is important to performance, as observed in experiments. Thus, the global-context normaliser is useful to avoid unfair weight assignment, while retaining the softmax layer.

The mathematical algorithm for the global context-normalised centrality routing is presented below. The primary differences between this and the routing procedure in [8] is the use of less bulky single GCNN-layers and, more importantly, the global-context normalisation layer. The routing algorithm is presented below.

### 3.2.1 The global context-normalised centrality routing algorithm

**Input:** $\left\{ f_i^l \,\middle|\, i \in \{0, ..., N_l - 1\},\ f_i^l : G \to R^{d^l} \right\}$

**Output:** $\left\{ f_j^{l+1} \,\middle|\, j \in \{0, ..., N_{l+1} - 1\},\ f_j^{l+1} : G \to R^{d^{l+1}} \right\}$

**Trainable functions:** $\left(\Psi_j^{l+1}, \star\right), 0 \leq j \leq N_{l+1} - 1$, $\star$ is the group equivariant convolution operator. $\Psi_j^{l+1}$ is indexed further by $p \in \{0, ..., d^{l+1} - 1\}$, where each $\Psi_j^{l+1,p} : G \to R$.

1. $S_{ijp}(g) = \left(f_i^l \star \Psi_j^{l+1,p}\right)$
2. $S^{l+1} = GlobalConNorm\left(S^{l+1}\right)$.
3. $c_{0j}^{l+1}(g), ..., c_{N_i^l - 1j}^{l+1}(g) = DegreeScore\left(S_{0j}^{l+1}(g), ..., S_{N_i^{l+1}j}^{l+1}(g)\right)$
4. $f_j^{l+1}(g) = Squash\left(f_j^{l+1}(g)\right) = \frac{||f_j^{l+1}(g)||}{||1 + f_j^{l+1}(g)||^2}$

**Procedure:** $DegreeScore\left(S_{0j}^{l+1}(g), ..., S_{0j}^{l+1}(g)\right)$

1. $A_{ik}^j(g) = \frac{S_{ij}^{l+1}(g) \cdot S_{kj}^{l+1}(g)}{||S_{ij}^{l+1}(g)|| \cdot ||S_{kj}^{l+1}(g)||}$

2. $Degree_i^j(g) = \Sigma_{k=0}^{N_l-1} A_{ik}^j(g)$

3. $c_{ij}^{l+1}(g) = \dfrac{\exp\exp\left(Degree_i^j(g)\right)}{\Sigma_{i=0}^{N_l-1}\exp\exp\left(Degree_i^j(g)\right)}$

**return** $c_{ij}^{l+1}(g)$

As described previously, the input capsule-types are first input to predictor GCNNs to obtain predictions for deeper capsule types. The predictions pass through the global context normalisation layer, denoted by GlobalConNorm, and then used to obtain centralities. The centralities are softmaxed to obtain routing-weights, which are then used in a weighted-summation to form deeper capsules from predictions. These deeper capsules are rescaled to have a norm between 0 and 1 using the squash operation of [5].

In the architectures we used, the ConvCaps layers were used sequentially with the number of capsules of the last ConvCaps layer being equal to the number of classes. This ensures that the RobustCaps model learns to represent the pose of the class objects in the last layer. Consequently, the hidden capsule layers can be thought of as learning to detect intermediate objects that are components of the class objects.

### 3.3 ProjCaps

After the final ConvCaps layer, the output gives the poses, across transformational states, of detected class objects. In the classification setting, we wish that only the most high scoring prediction be used. In order to obtain scores from the vector capsules, many capsule networks such as DeepCaps, the model in [5], and the model in [7] use the 2-norm of the capsules as the score. In our work, we propose an alternative that works well for RobustCaps.

We project each capsule to a scalar using a GCNN that is shared among all the class capsule types. Thus, we use a learnable mechanism for classification instead of using the 2-norm. The sharing of the GCNNs allows for parameter-efficiency as certain classes such as CIFAR-100 have a relatively large number of classes.

Following the projection, the higher scalar value across all transformational states for a class capsule type is taken as the score for that capsule type. The prediction for RobustCaps is taken as the index of the class capsule type that has the highest score.

## 4. Experiments and results

A challenging task for models, that is close to the real-world setting, is obtaining high performance while training and testing in the presence of significant geometric transformations of the input. Thus, we conduct experiments on transformed image classification on three datasets, namely CIFAR-10, FashionMNIST, and CIFAR-100. We explain the setting of these experiments.

### 4.1 Transformed image classification

Given a dataset of images in a classification setting, transforming the train or test images by geometric transformations does not, in general, change the category of the image. It however changes the configuration of the pixels of the image. The effect and purpose of transformations of the train set and test set are different.

Test time transformations can be used as a mechanism to showcase the transformation robustness of models. Models that show good performance on transformed datasets can be thought of as being more robust to transformations. Train time transformations, on the other hand, can be used as a means of checking the ability of a model to learn from transformations. Together, building a train and test system where train and test time transformations both are used allows the checking of a model's transformation-robustness and ability to learn from transformed data.

A similar setup was explored in [8], where a variety of train and test time transformations were used. In our work, we study the most challenging part of this setup and use this to investigate the models' behaviour. Concretely, we modify a given dataset by randomly transforming its train set by translating by up to 2 pixels in the x and y directions, and then by rotating the image by a randomly chosen angle in the range of (-180°, 180°). We represent this by a tuple of (2, 180°).

More generally, the transformation extents for a dataset can be written as (tr, rot), where tr gives the translational extent and rot given the rotational extent. In order to study the robustness of models, we create five tests from the original test set. We retain the test set as the first set, and then create four more by using a translational extent of 2 pixels, and rotational extents of 30°, 60°, 90°, and 180°. The train and test transformational extents are given in Table 1.

### 4.2 Description of the datasets

With the transformed classification set up explained above, we present a description of the datasets used in our experiments.

*FashionMNIST* The FashionMNIST dataset [13] consists of a train and a test set, each consisting of grayscale images of clothing items. The train set consists of 60,000 28×28 images of clothing items, belonging to ten categories. Each category has 6000 images in the train dataset. The test set consists of 10,000 28×28 images of clothing. Each category has 1000 images in the test dataset. FashionMNIST is a reasonably challenging dataset for mid-sized deep neural networks, and obtaining high accuracies on the transformed test datasets showcases the performance of models.

*CIFAR-10* The CIFAR-10 dataset [14] consists of a train and test dataset, each consisting of colour images of general objects. The train set consists of 50,000 32×32 images of general objects belonging to 10 categories. Each category has 5000 images in the train dataset. The test set consists of 10,000 32×32 images. Each category has 1000 images in the test dataset. CIFAR-10 is more challenging than FashionMNIST as it introduces aspects such as variable backgrounds and differently posed objects.

*CIFAR-100* The CIFAR-100 dataset [14] consists of a train and test dataset, each consisting of colour images of general objects. The train set consists of 50,000 32×32 images of general objects belonging to 100 categories. Each category has 500 images in the train dataset. The test set consists of 10,000 32×32 images. Each category has 100 images in the test dataset. CIFAR-100 is more challenging than FashionMNIST and CIFAR-10 as it introduces a larger category set while reducing the number of images per class, along with other challenging aspects variations in background and differently posed objects.

All of the above datasets were used to create a train dataset and 5 test datasets as described previously. During train time, we also augmented the train dataset in a few experiments by random crops and randomly chosen horizontal flips. This augmentation is denoted by the suffix -aug in the Tables. These augmentations are simple augmentations, and used frequently in the literature.

### 4.3 Description of baselines

In order to present a fair and diverse comparison for the robustness of RobustCaps, we have considered several capsule baselines. These models are as follows: CapsNet [5], EMCaps [6], GCAPS [9], DeepCaps [11], SOVNET [7]. We also trained and tested the following CNN baselines: ResNet-18 [15], ResNet-34, and their group-equivariant versions defined on the p4m group [16], and denoted by GResNet-18 and GResNet-34.

The performance on capsule baselines on the three datasets have been reported in [7] - these are mentioned for the sake of comparison in Table 2, Table 3, and Table 4. All of the above models have been tested on FashionMNIST and CIFAR-10. For CIFAR-100, we tested only the best performing baselines as CIFAR-100 is challenging for most models.

Table 1 The translational and rotational extents to generate the datasets.

| Name of dataset | Translational | Rotational | Representation |
|---|---|---|---|
| Train dataset | 2 | 180° | (2, 180°) |
| First test dataset | 0 | 0° | (0, 0°) |
| Second test dataset | 2 | 30° | (2, 30°) |
| Third test dataset | 2 | 60° | (2, 60°) |
| Fourth test dataset | 2 | 90° | (2, 90°) |
| Fifth test dataset | 2 | 180° | (2, 180°) |

### 4.4 Description of the RobustCaps architecture

The RobustCaps architecture that we used for our experiments involves 7 residual blocks for the PreCaps layers, a single PrimaryCaps layer that uses 32 capsule-types of 16 dimensions, 4 ConvCaps layers that use 32 capsule-types of 16 dimensions, except for the last layer that uses capsule-types equal to the number of classes. All the GCNN layers were defined over the p4 group that consists of translations composed with rotations of multiples of 90°. The results comparing the classification accuracies on the test sets are given in Table 2, Table 3, and Table 4.

### 4.5 Training of RobustCaps

All the RobustCaps models were trained using 3 Nvidia GeForce RTX 2080 ti GPUs. The models were trained using cross entropy loss with the AdamW optimiser [17] and a OneCycleLR [18] scheduler. All these models were trained for 150 epochs. The code was written in pytorch.

### 4.6 Results of our experiments

#### 4.6.1 Comparison with capsule networks

First, we see that RobustCaps outperforms all the capsule baselines on each test set, including the equivariant models GCAPS and SOVNET. This is because of the equivariance and improved model-structure of RobustCaps. In particular, the low performance of GCAPS on CIFAR-10 and CIFAR-100 suggests that equivariance is not enough to attain good performance - a sufficiently strong model structure is also necessary. The performance of RobustCaps is evidence for this.

On FashionMNIST, we see that all capsule baselines achieve over 80% performance. This is because the dataset has a single background, and the complexity of the dataset is relatively low. Nevertheless, there is a significant gap in the performance between RobustCaps and most capsule networks. The models closest to the performance of RobustCaps is SOVNET and SOVNET-aug. Even these models showcase a gap of around 0.1-0.4% gap. In particular, we see that the gap is largest on the test set transformed by (2, 180°), This showcases the improved transformation robustness of RobustCaps.

While RobustCaps performs better than other capsule network models on FashionMNIST, the dataset itself does not fully bring out a challenge to the models. As mentioned, this is because of its simpler objects and shared background across all data. CIFAR-10 and CIFAR-100 differ in this. They showcase more complex objects and background information. This makes them harder to classify.

We see that RobustCaps beats the closest capsule model by almost 10% on all test sets for CIFAR-10. This showcases the need for an improved architecture along with equivariance. On CIFAR-100, we evaluated only the best-performing model from before, that is SOVNET. SOVNET performs poorly on CIFAR-100, achieving low accuracies on all test sets. RobustCaps achieves much higher than SOVNET on all test sets.

#### 4.3.2 Comparison with CNN models

We see that the residual networks ResNet18, ResNet34, GResNet18, and GResNet34 perform well across all datasets. In particular, the equivariant resnets perform well and display great transformation-robustness. Thus, these are strong baselines for comparison. RobustCaps outperforms these baselines on all the test sets of all the datasets, indicating that it is both high-performing and capable of robustness.

The results showcasing the improved performance of RobsuCaps over resnets is important for capsule network research. This is due to the fact that most capsule network models do not perform as well as CNN models on classification tasks. By showcasing improvements over even equivariant CNN models, our work can renew interest in capsule research.

Table 2 The accuracies of various models on transformed classification for FashionMNIST. The training images have been translated by pixels up to 2 pixels and rotated by a random angle between (-180°, 180°). The results of the models on 5 test datasets have been given. **Our model achieves the best results on all the 5 test datasets.**

| Method | (0, 0°) | (2, 30°) | (0, 60°) | (2, 90°) | (2, 180°) |
|---|---|---|---|---|---|
| CapsNet | 86.90% | 84.94% | 84.93% | 84.75% | 84.72% |
| EMCaps | 82.99% | 82.67% | 82.18% | 82.32% | 82.18% |
| GCaps | 80.65% | 79.66% | 79.46% | 79.47% | 79.37% |
| DeepCaps | 92.07% | 91.71% | 91.70% | 91.76% | 91.66% |
| SOVNET | 94.11% | 93.77% | 93.56% | 93.57% | 93.60% |
| SOVNET-aug | 94.21% | 93.58% | 93.46% | 93.57% | 93.61% |
| ResNet-18-aug | 94.21% | 93.55% | 93.24% | 93.30% | 93.45% |
| ResNet-34-aug | 94.38% | 93.75% | 93.78% | 93.78% | 93.73% |
| GResNet-18-aug | 93.63% | 93.38% | 93.32% | 93.31% | 93.35% |
| GResNet-34-aug | 93.22% | 92.71% | 93.08% | 93.01% | 92.81% |
| **RobustCaps-aug (ours)** | **94.33%** | **93.98%** | **93.87%** | **94.01%** | **94.07%** |

Table 3 The accuracies of various models on transformed classification for CIFAR-10. The training images have been translated by pixels up to 2 pixels and rotated by a random angle between (-180°, 180°). The results of the models on 5 test datasets have been given. **Our model achieves the best results on all the 5 test datasets.**

| Method | (0, 0°) | (2, 30°) | (0, 60°) | (2, 90°) | (2, 180°) |
|---|---|---|---|---|---|
| CapsNet | 61.08% | 59.53% | 60.04% | 59.85% | 59.90% |
| EMCaps | 57.57% | 55.89% | 56.85% | 56.35% | 55.20% |
| GCaps | 39.09% | 41.03% | 41.43% | 41.25% | 41.08% |
| DeepCaps | 81.12% | 80.81% | 80.64% | 81.05% | 80.92% |
| SOVNET | 82.50% | 81.80% | 81.78% | 81.95% | 81.82% |
| SOVNET-aug | 80.14% | 79.64% | 79.94% | 79.99% | 79.65% |
| ResNet-18-aug | 78.84% | 79.28% | 79.72% | 79.60% | 78.95% |
| ResNet-34-aug | 81.27% | 81.15% | 81.44% | 81.60% | 81.65% |
| GResNet-18-aug | 89.88% | 89.46% | 89.33% | 89.54% | 89.41% |
| GResNet-34-aug | 89.12% | 89.02% | 89.18% | 88.85% | 89.10% |
| **RobustCaps-aug (ours)** | **92.01%** | **91.44%** | **91.39%** | **91.25%** | **91.36%** |

Table 4 The accuracies of various models on transformed classification for CIFAR-100. The training images have been translated by pixels up to 2 pixels and rotated by a random angle between (-180°, 180°). The results of the models on 5 test datasets have been given. **Our model achieves the best results on all the 5 test datasets.**

| Method | (0, 0°) | (2, 30°) | (0, 60°) | (2, 90°) | (2, 180°) |
|---|---|---|---|---|---|
| SOVNET | 40.38% | 39.74% | 39.84% | 39.76% | 39.77% |
| SOVNET-aug | 40.38% | 39.82% | 39.69% | 39.78% | 39.99% |
| ResNet-18-aug | 50.03% | 50.56% | 51.15% | 51.00% | 51.14% |
| ResNet-34-aug | 51.40% | 51.86% | 51.49% | 51.93% | 52.11% |
| GResNet-18-aug | 64.22% | 64.19% | 63.86% | 63.89% | 63.38% |
| GResNet-34-aug | 66.12% | 65.90% | 65.66% | 65.56% | 65.92% |
| **RobustCaps-aug (ours)** | **67.60%** | **67.03%** | **67.40%** | **66.96%** | **67.18%** |

### 4.4 Summary of results

We see that our RobustCaps model outperforms all baselines on all test sets on the challenging transformed classification task. Specifically, the improved performance seen against CNN baselines shows that capsule networks have the potential to achieve good results, and can renew interest in capsule research.

### 4.5 Ablation studies

In this section, we present some ablation studies to showcase the importance of using the PreCaps and ProjCaps layers. Specifically, we train and test certain RobustCaps models on the transformed classification on CIFAR-10 that do not use the PreCaps and ProjCaps layers of the model. A dip in performance is indicative of the importance of these layers that we used.

Table 5 The accuracies of three RobustCaps models on transformed tests of CIFAR-10. NoPreCaps uses a simple convolutional pre-capsule layer. NoProjCaps uses the 2-norms of final layer capsules for predictions. Note the importance of these subnetworks, as their inclusion improves the performance of the model.

| Method | (0, 0°) | (2, 30°) | (2, 60°) | (2, 90°) | (2, 180°) |
|---|---|---|---|---|---|
| NoPreCaps | 75.09% | , 74.67% | 74.94% | 74.76% | 74.95% |
| NoProjCaps | 90.30% | 89.63% | 89.50% | 89.62% | 89.53% |
| RobustCaps | 92.01% | 91.44% | 91.39% | 91.25% | 91.36% |

### 4.5.1 Removal of PreCaps

We trained and tested a modified RobustCaps model that uses a simple convolutional pre-capsule network instead of PreCaps. This network uses convolutions with batchnorm and relu. The model was trained and tested on the transformed versions of CIFAR-10 just as the original RobustCaps model. The performance of this model and the original RobustCaps model is given in Table 5. This model is given as NoPreCaps.

We see that the performance suffers a significant decrease if PreCaps is substituted for a simple convolutional network. This is because the capsule intuition requires objects at a sufficiently high semantic level. Using deeper PreCaps layers allows for the detection of such patterns.

### 4.5.2 Removal of ProjCaps

We train and test a RobustCaps model where, instead of using the ProjCaps subnetwork, the predictions are based on the 2-norms of the capsules of the last layer. This is the same approach used in most capsule networks. The model was trained on the CIFAR-10 dataset with the same transformations as RobustCaps. The performance of this model and the original RobustCaps model is given in Table 5. This model is given as NoProjCaps.

We see that the performance suffers a decrease. This, while not as high as that of NoPreCaps, is still high enough to warrant as significant.

## 6. Conclusion

In conclusion, we have presented a group-equivariant model for transformation-robust capsule networks, termed RobustCaps. RobustCaps uses GCNNs in a capsule framework, allowing for the learning of transformation-invariant relationships, while ensuring that the routing-by-agreement principle of capsule networks is not affected by factors such as scale of activations.

Specifically, our model improves upon capsule network models by using depth-specific subnetworks that use equivariant mechanisms to build capsules. Our experiments show that RobustCaps outperforms several strong baselines. This establishes the efficacy of part-whole learning in an equivariant framework.

We wish to investigate the use of such a model in more complex tasks where object-relationships are more explicit and complex. Models along these lines can lead to better interpretability.

## Acknowledgments

The authors dedicate this work to the founder chancellor of Sri Sathya Sai Institute of Higher Learning, Bhagavan Sri Sathya Sai Baba.